\documentclass{article}

\usepackage{comment}
\usepackage{gensymb}
\usepackage{graphics} 
\usepackage{epsfig} 
\usepackage{mathptmx} 
\usepackage{times} 
\usepackage{amsmath} 
\usepackage{amssymb}  
\usepackage{xcolor}
\usepackage{subcaption}
\usepackage{graphicx}
\usepackage{wrapfig}
\usepackage{caption}
\usepackage{soul}
\usepackage{hyperref}
\usepackage[title]{appendix}
\hypersetup{
    colorlinks=true, 
    linkcolor=blue, 
    filecolor=black, 
    urlcolor=blue, 
}

\DeclareMathAlphabet{\mathcal}{OMS}{cmsy}{m}{n}

\usepackage[final]{corl_2020_arxiv}

\title{S3K: Self-Supervised Semantic Keypoints for Robotic Manipulation via Multi-View Consistency}

\newcommand{\ie}{\textit{i}.\textit{e}. }
\newcommand{\eg}{\textit{e}.\textit{g}. }
\newcommand\sref{Section~\ref}
\newcommand\eref{Eq.~\ref}
\newcommand\fref{Fig.~\ref}
\newcommand\tref{Table~\ref}

\newcommand\eqshrink{\vspace{-0.4cm}}
\newcommand\eqshrinkb{\vspace{-0.3cm}}

\author{
  Mel Vecerik\\
  University College London, DeepMind \\ 
  \texttt{mel.vecerik.18@ucl.ac.uk}, \texttt{vec@google.com} \\
  \And
  Jean-Baptiste Regli \\
  DeepMind \\
  \texttt{jbregli@google.com} \\
  \And
  Oleg Sushkov\\
  DeepMind \\
  \texttt{sushkov@google.com} \\
  \And
  David Barker\\
  DeepMind \\
  \texttt{davebarker@google.com} \\
  \And
  Rugile Pevceviciute \\
  DeepMind \\
  \texttt{rugile@google.com} \\
  \And
  Thomas Roth{\"o}rl \\
  DeepMind \\
  \texttt{tcr@google.com} \\
  \And
  Christopher Schuster\\
  DeepMind \\
  \texttt{cschuster@google.com} \\
  \AND
  Raia Hadsell \\
  DeepMind \\
  \texttt{raia@google.com} \\
  \And
  Lourdes Agapito \\
  University College London \\
  \texttt{l.agapito@cs.ucl.ac.uk} \\
  \And
  Jonathan Scholz \\
  DeepMind \\
  \texttt{jscholz@google.com} \\
}

\begin{document}
\maketitle


\begin{abstract}

A robot's ability to act is fundamentally constrained by what it can perceive.
Many existing approaches to visual representation learning utilize general-purpose training criteria, \eg image reconstruction, smoothness in latent space, or usefulness for control, or else make use of large datasets annotated with specific features (bounding boxes, segmentations, etc.).
However, both approaches often struggle to capture the fine-detail required for precision tasks on specific objects, \eg grasping and mating a plug and socket.
We argue that these difficulties arise from a lack of \textit{geometric structure} in these models.
In this work we advocate \emph{semantic 3D keypoints} as a visual representation, and present a self-supervised training objective that can allow instance or category-level keypoints to be trained to 1-5 millimeter-accuracy with minimal supervision.
Furthermore, unlike local texture-based approaches, our model integrates contextual information from a large area and is therefore robust to occlusion, noise, and lack of discernible texture.
We demonstrate that this ability to locate \emph{semantic keypoints} enables high level scripting of human understandable behaviours.
Finally we show that these keypoints provide a good way to define reward functions for reinforcement learning and are a good representation for training agents.

\end{abstract}



\begin{wrapfigure}{r}{2.5cm}
    \vspace{-2.15cm}
    \centering
    \includegraphics[width=\linewidth]{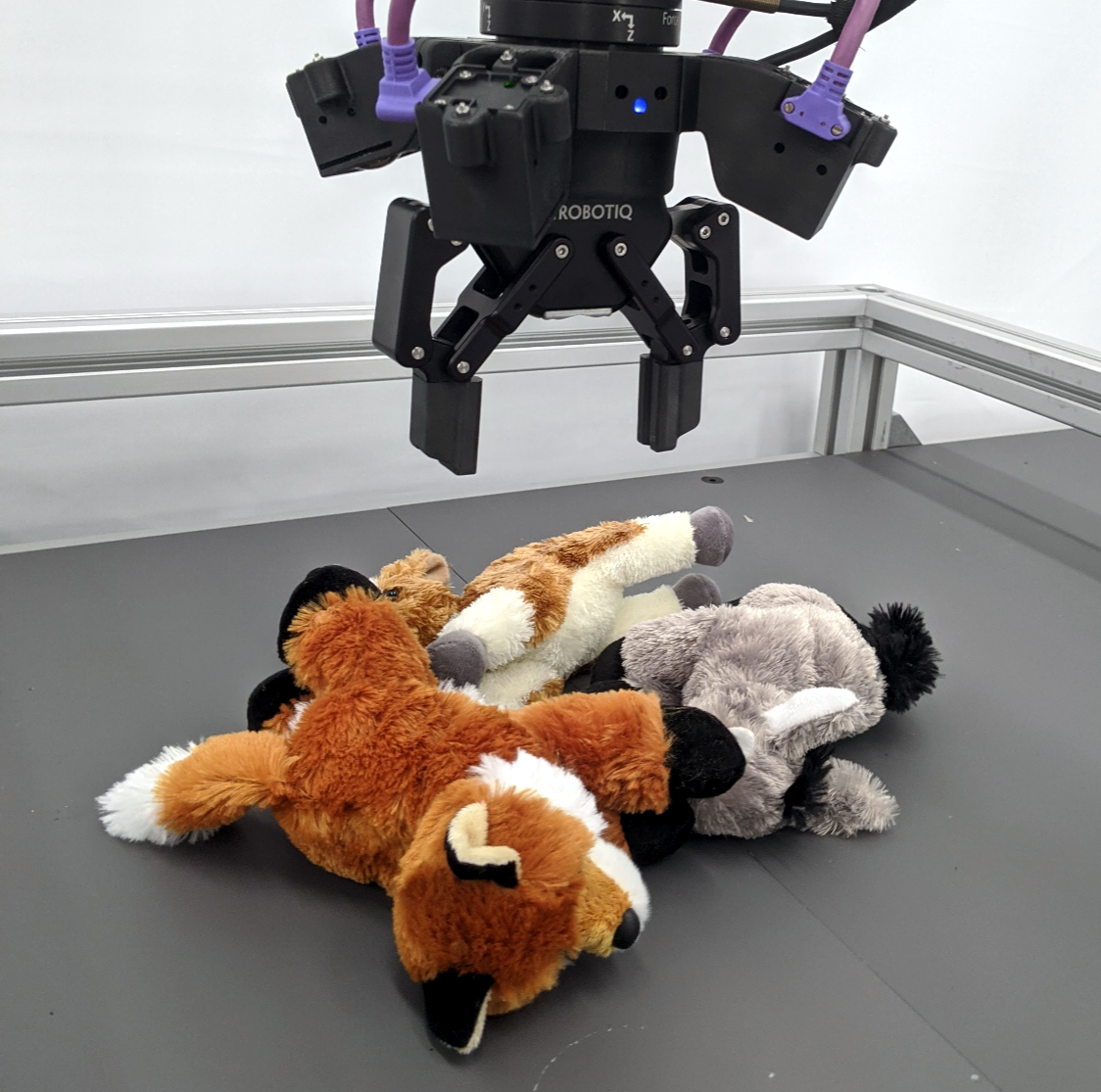}
    \includegraphics[width=\linewidth]{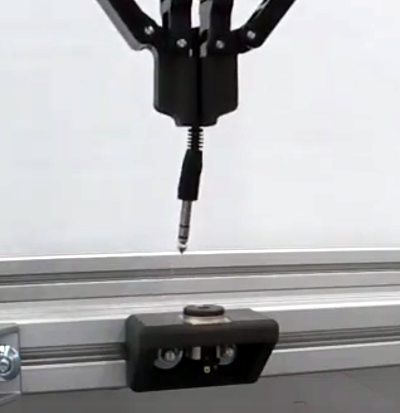}
    \caption{Fox grasp and cable insertion tasks.}
    \label{fig:image:cable_foxie_scenes}
    \vspace{-1.0cm}
\end{wrapfigure}


\section{Introduction}

It is well understood that manipulating objects precisely requires (at least) an implicit understanding of the 3D relationships between the robot and objects in space.
Most existing visual representations achieve this either by full supervision \cite{xiang2017posecnn, rad2017bb8, peng2019pvnet, liu2020keypose}, dense self-supervision \cite{rhodin2018unsupervised, tang2019selfsupervised}, or else lack 3D structure entirely \cite{Vecerk2019APA, finn2016deep, yen2020learning, Oord2018CPC, kingma2013autoencoding, redmon2016yolo9000}.
In this paper we explore a middle ground: our goal is to learn an explicit 3D representation that captures \emph{task-relevant} object geometry, without requiring depth-sensing or full supervision.
We argue that these requirements are critical in order to achieve generalizable manipulation behavior in an inherently 3D world, while being practical to train.

Our approach utilizes multiple camera views to train a model to extract keypoints located on relevant object parts.
This representation offers two key benefits: 1) the use of multi-view geometry constrains the hypothesis space to geometrically meaningful features, and 2) it provides an intuitive way for humans to tell robots what to attend to, \eg by clicking on an image.

However, this approach raises the question of how to obtain sufficient annotations without a tiresome labeling process.
S3K is a novel semi-supervised approach which allows us to use a small number of 2D annotations to specify what to focus on, along with a large corpus of unlabeled images to generalize to a wide range of conditions.

Our central finding is that the self-supervised component of the model is capable of extracting 3D structure from data even with an extreme scarcity of ground-truth labels.
This allows a keypoint detector to be trained to a high accuracy using a surprisingly small number of annotated images. 
In addition, our approach naturally permits keypoints not to be tied to surface points -- they can be inside objects or correspond to no physical location at all as long as they are semantically and geometrically consistent.
We evaluate this model across a set of simulated and real-robot manipulation tasks, involving both rigid and deformable objects.
Videos and further materials available at \href{https://sites.google.com/view/2020-s3k/home}{\color{blue}{sites.google.com/view/2020-s3k}}.

\section{Related work}
\label{sec:related_work}

\begin{figure}[!bp]
    \begin{minipage}{0.48\textwidth}
        \centering
        \begin{minipage}{0.5\textwidth}
            \includegraphics[width=\textwidth, height=\textwidth]{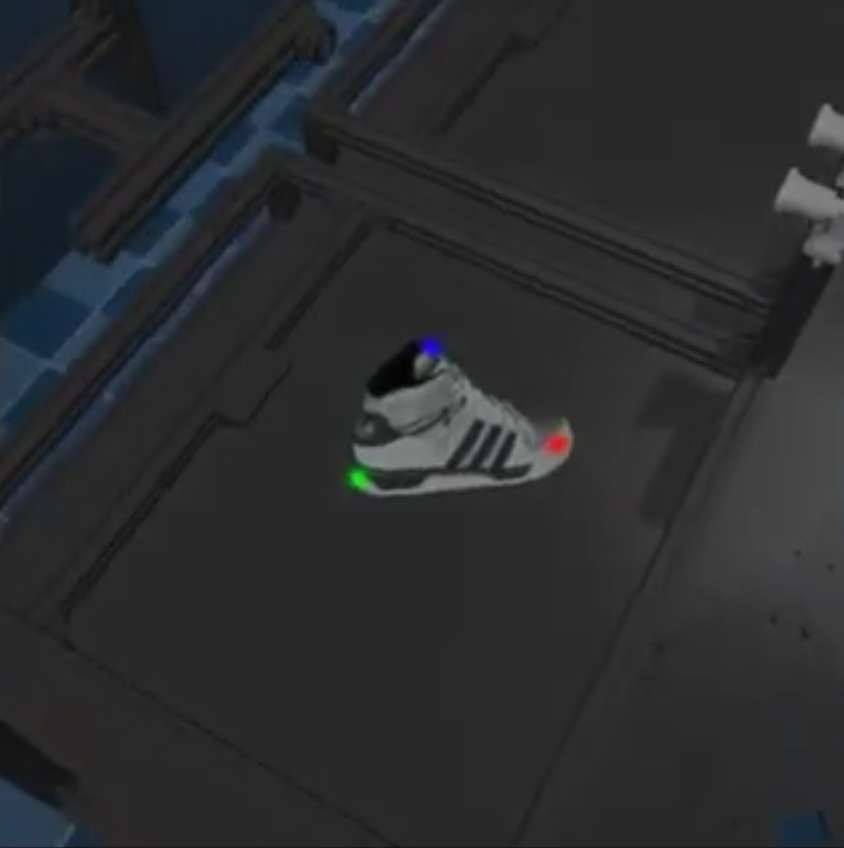}
        \end{minipage}%
        \begin{minipage}{0.5\textwidth}
            \includegraphics[width=\textwidth, height=\textwidth]{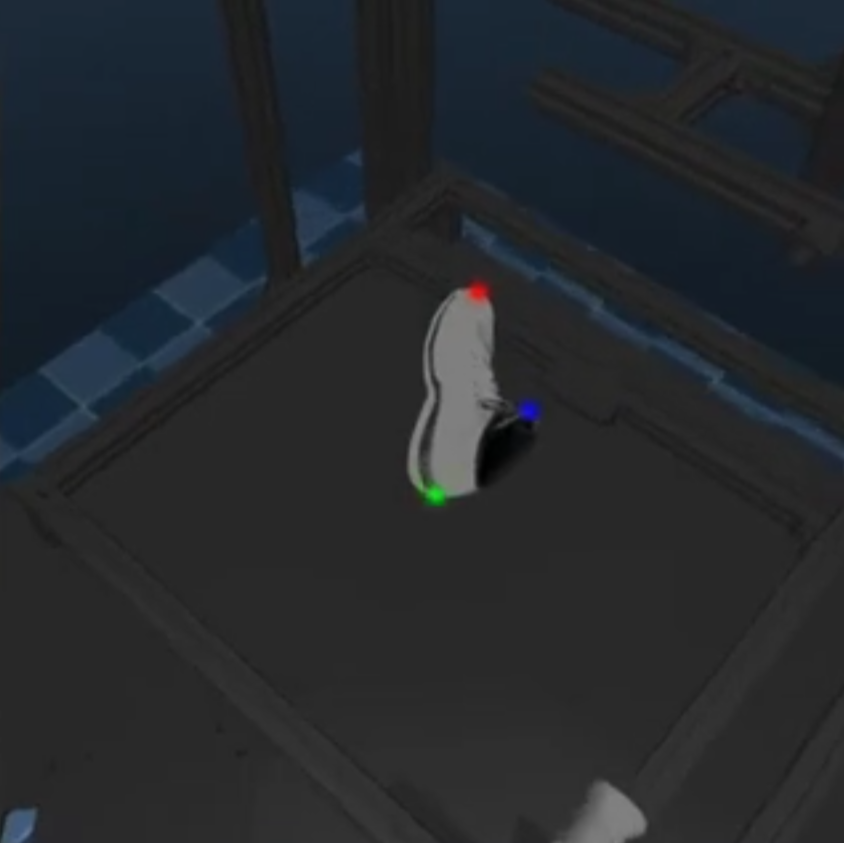}
        \end{minipage}%
        \caption{Exploring the generalization ability of the keypoint model.}
        \label{fig:img:shoe}
    \end{minipage}
    \hfill
    \centering
    \begin{minipage}{0.48\textwidth}
        \begin{minipage}{0.5\textwidth}
            \includegraphics[width=\textwidth, trim=50 21 10 12, clip]{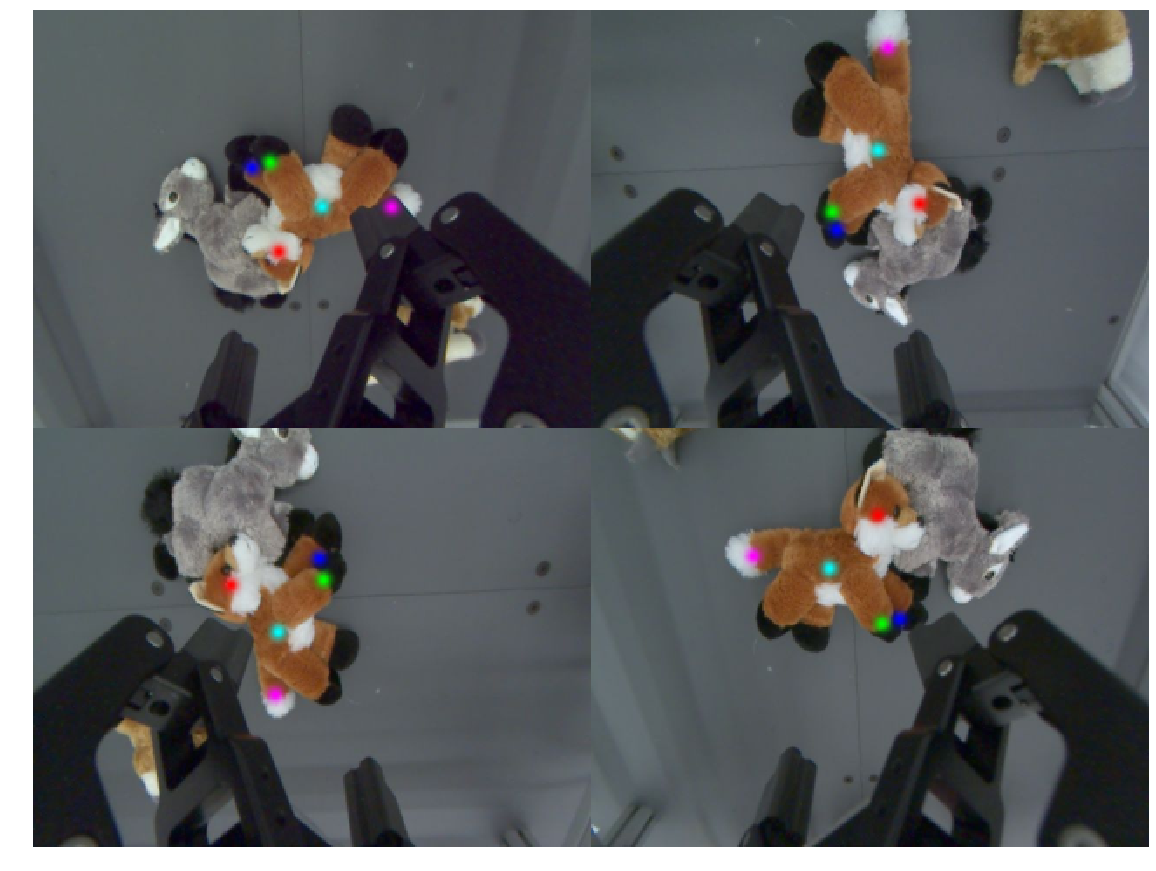}
        \end{minipage}%
        \begin{minipage}{0.5\textwidth}
            \includegraphics[width=\textwidth, trim=50 21 10 12, clip]{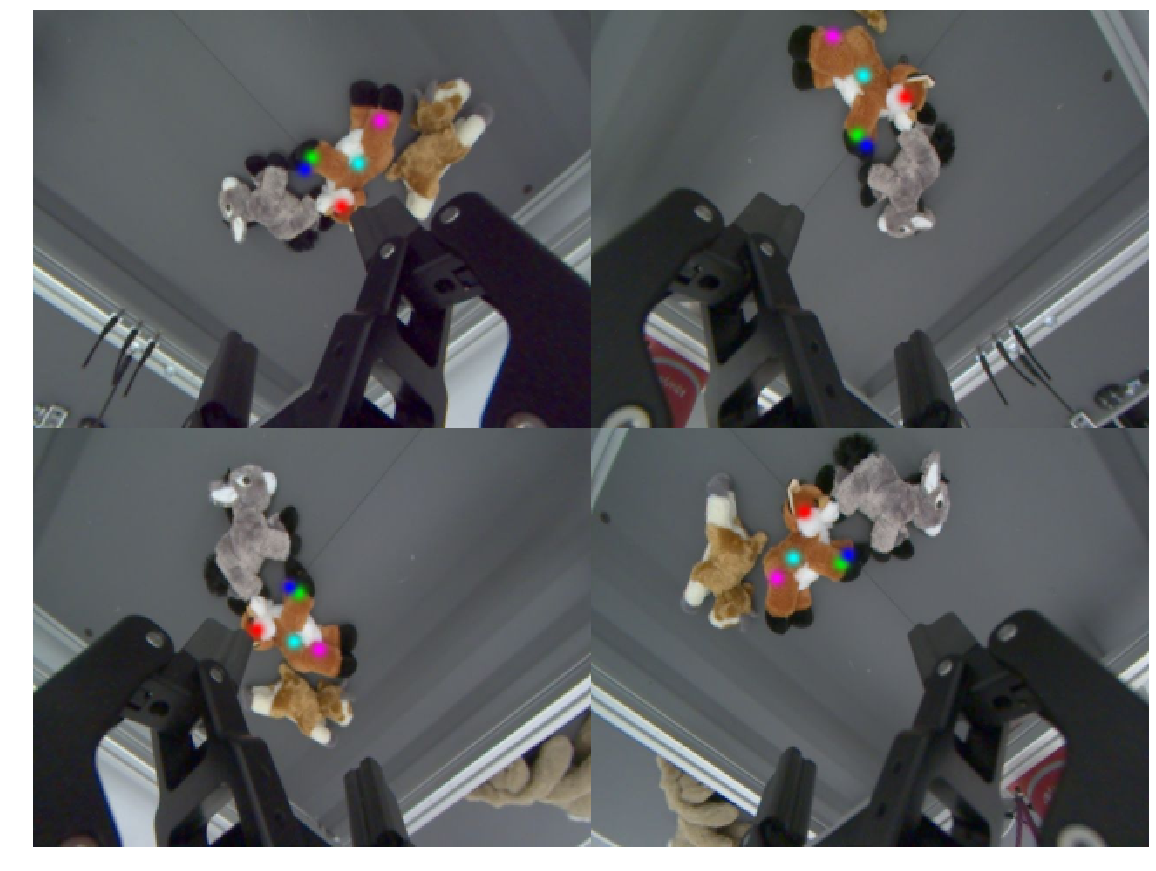}
        \end{minipage}%
        \caption{Keypoint detections of unseen configurations of the plush objects.}
        \label{fig:image:foxie}
    \end{minipage}
\end{figure}

The overarching goal of this paper is to enable the creation of new behaviours through better representations.
\citet{finn2016deep} showed that policies can be trained on top of a spatial-softmax operation which allows a neural network to extract 2D image \emph{coordinates} from convolutional feature maps.
This operator, however offers no explicit reasoning in 3D space and in practice feature maps can be noisy and waste representational capacity on task-irrelevant detail, especially in non-planar tasks.
This effect can be reduced \eg by using triplet or contrastive losses \cite{triplet_JMLR:v11:chechik10a}, or by using self-supervision across multiple input modalities \cite{Lee2019VisionAndTouch}.
TCN \cite{TCN2017} achieves this in a multi-view setting similar to ours, but does not exploit camera geometry.
Recently, several works use invariance to cropping as a way to constrain representations  \cite{laskin_lee2020rad, laskin_srinivas2020curl, Kostrikov2020ImageAI}.
However, all of these representations lack 3D awareness and are therefore hard to interpret within the setting of a robot controller.

In 3-dimensions, keypoints are a foundational and common representation that can be used to extract higher-level object features such as 6D-pose \cite{peng2019pvnet, liu2020keypose}.
Alternatively they can be used directly for robot manipulation as in \cite{Manuelli2019kPAMKA} which defines an optimization problem (\eg shoe reorientation) in terms of 3D keypoints.
\cite{liu2020keypose} showed that keypoints can significantly improve 6-DOF tracking performance on transparent objects, which is known to be a challenging problem for both template and direct-regression methods \cite{li2018deepim}.
\cite{Manuelli2019kPAMKA} and \cite{liu2020keypose} also described a geometry-based mechanism to allow label propagation across time, which we employed in our grasp experiments below.
However, this treatment requires the scene to be static, which precludes training from dynamic interactions.

Other related methods utilize keypoints internally to obtain 6D pose estimates, but do not require explicit supervision on the keypoints \cite{tang2019selfsupervised, cieslewski2018matching, NIPS2018_7476}.
Representations constructed for camera localization are biased to discard moving objects which makes them unsuitable for most manipulation tasks. 
A more robotics-relevant approach was shown in \cite{jakab2018unsupervised} which uses reconstructions within episodes to learn a keypoint detector in an unsupervised way.
However, rich and moving backgrounds quickly use up model capacity which limits their ability to precisely represent small objects or specific object parts.

Keypoints are also very common in human and hand pose tracking \cite{CHEN2020102897}.
\cite{rhodin2018unsupervised} enforces a 3D-aware latent representation by predicting an image from one viewpoint given an image from another. 
However, the image reconstruction loss makes it dependent on high quality background subtraction.
A different approach was demonstrated by \cite{rhodin2018skiing} where consistency across multiple cameras is used to predict dynamic poses of skiers.
This work is similar to ours however they do not use calibrated cameras, which restricts them to predicting \textit{normalized} poses. 
This is an important limitation for us because in our tasks the absolute scale is important.

Lastly, the goal of our work is also closely related to \cite{florencemanuelli2018dense} where a mapping is learned from camera images to a per-pixel dense embedding.
This is done by considering pixels which represent the same spatial location from different views and considering their embeddings as positives in a contrastive loss.
3D keypoints can be constructed by locating a specific embedding in pixel space and combining it with depth information.
Further in \cite{Florence2020densenets_visuomotorpolicy} demonstrates that these methods can work from dynamic depth maps by using simultaneous image acquisitions from different views, instead of different time-points.
This is the same approach we pursue here, but we triangulate depth from multi-view geometry rather than requiring a calibrated depth-sensor.
A limiting factor of these approaches is that they rely on having quality depth maps to create the image correspondences and most current depth sensors struggle to provide good data for thin or highly reflective objects.
In addition, depth-sensors only provide correspondences on the object surface, unlike our method and \cite{Manuelli2019kPAMKA}, which we believe is important in order to handle occlusions or arbitrary rotations.

\section{Method}
\label{sec:method}

\begin{figure}[]
  \centering
\includegraphics[width=\textwidth]{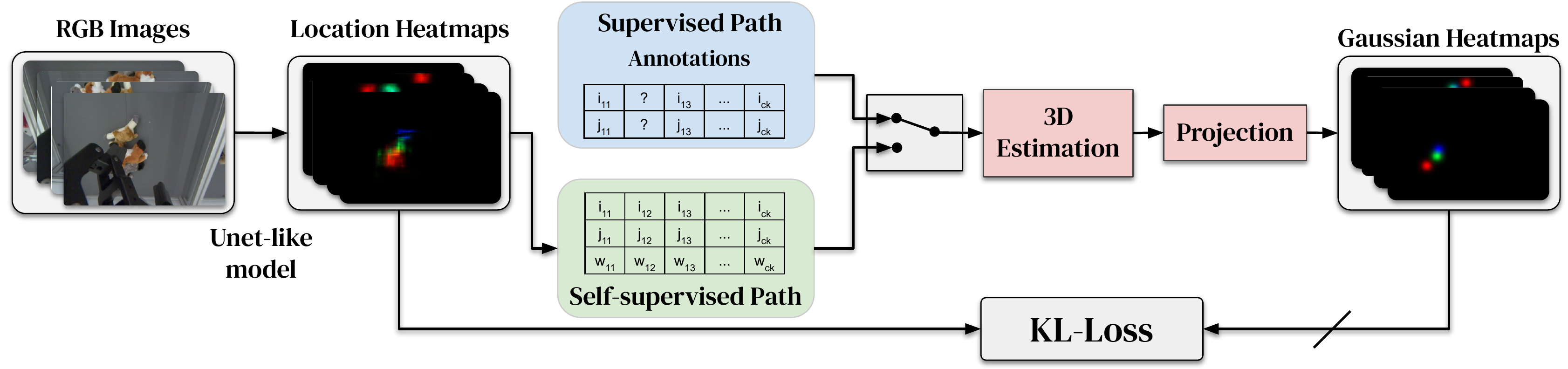}
    \caption{
    Diagram of the learning setup showing the supervised and the self-supervised paths.
    The model predicts location heatmaps for each camera and keypoint.
    To learn from annotations we need the keypoint to be labelled in at least 2 out of 4 cameras.
    Otherwise we use the coordinates and uncertainties from the model to estimate the keypoint locations instead.
    These 3D locations are then projected back to the camera frames and provide a learning target for the model via a KL-loss.
    }
  \label{fig:diag:semi-supervised}
  \vspace{-0.5cm}
\end{figure}

There are two core components of S3K: (1) a neural-network which is trained to detect 2D keypoints for all individual scene views, and (2) a 3D estimation layer which consumes the 2D predictions and generates an estimate of the 3D location of each keypoint.
At a high-level, the network is trained to achieve a spatial consensus among the 2D predictions by projecting the 3D estimate \textit{back} to each camera view and penalizing divergence.
This structure allows the network to be trained from a mixture of labelled and unlabelled data -- labels are only required to ground the model's ability to target specific points, beyond which consensus is sufficient to drive further generalization.
In this section we detail the operation of both training pathways, and a simple heuristic for handling missing or uncertain predictions.
\fref{fig:diag:semi-supervised} depicts the structure of these components.

\subsection{Supervised keypoint training}

In the supervised regime we have access to 2D labels of a single point in space from multiple perspectives.
Given the camera calibration we can compute the 3D location $\mathbf{x}_k$ of all $K$ keypoints even when annotations from all cameras are not available.
Similarly, we can project any 3D location back onto the image planes.
We write the image coordinates of the back-projected keypoint $k$ into the camera image $c$ as $i_{ck}$, $j_{ck}$ and we construct a Gaussian image centered at this location with width $\sigma$.
This creates a heat-map of the following form: $H_{ck}(ij) \propto \exp \left ( (-(i_{ck} - i)^2 - (i_{ck} - j)^2)/(2 \sigma^2) \right) $ where $i$ and $j$ are coordinates in the heatmap.

To define the loss, we write the images as $I_c$, the model as $f$, and the model parameters as $\theta$.
Next we obtain the model's prediction $P_{ck} = \mathrm{softmax}(f(I_c, \theta)_k)$ where the $\mathrm{softmax}$ is performed separately for each keypoint and image.
We emphasize that the \emph{same} network is applied to each viewpoint to obtain the heatmaps for all keypoints. 
In the loss, we sum the $D_\mathrm{KL}$ terms between the heat-map $H_{ck}(ij)$ and the model prediction $P_{ck}$ for all $C$ camera images, and all $K$ keypoints.
From this we can write the final supervised loss as:

\eqshrink
\begin{equation}
    \mathcal{L}_{sup} = \sum_{c}^{C} \sum_{k}^{K} D_\mathrm{KL} \left (H_{ck} \middle|\middle| softmax(f(I_c, \theta)_{k}) \right )
\end{equation}
\eqshrinkb

\subsection{Self-supervised keypoint training}
In the self-supervised case we do not have explicit labels, but we do assume access to multiple simultaneous images of the scene.
Note that in the supervised case we did not directly train on the 2D annotations, but rather estimated a 3D location and re-projected this hypothesis to each camera to obtain a target heatmap.
The same procedure can be followed with any 3D location, so we can use a bootstrapped estimate rather than an annotation.
This allows the model to self-supervise from its own predictions in a 3D-consistent way.
This corresponds to the bottom pathway in \fref{fig:diag:semi-supervised}.

To estimate the 3D location we generate the heat-map predictions  $P_{ck} = \mathrm{softmax}(f(I_c, \theta)_k)$ as in the supervised case.
The mean of $P_{ck}$ corresponds to an image position, from which a direction can be computed via calibration similarly to \cite{liu2020keypose, Hodan2020EPOSE6}.
This creates a ray for each camera in the direction of its estimate.
Assuming that every keypoint appears on the scene exactly once, we define our keypoint position estimate $\widetilde{\mathbf{x}}_k$, as a point for which the sum of squares of distances from these rays is minimized.
This defines a quadratic cost for $\widetilde{\mathbf{x}}_k$ and therefore can be analytically solved (\eref{eq:weighted_unsup_solution}), 

\label{sec:quadratic_magic}

\subsubsection{Detection weight}

It is not always desirable to incorporate all predictions from all cameras, e.g. due to certain keypoints dropping out of view or becoming occluded.
Additionally, it is often desirable to put a higher weight on the camera views with the more confident predictions.
We explored several explicit ways to predict the confidence from the network, but found that the most effective way was to look at the variance $var_{ck}$ of the prediction heatmap $P_{ck}$ itself.
We defined the following confidence measure to weigh the importance of each heatmap:
$w_{ck} = \mathrm{sigm} (3 \mathrm{tanh} (5 (1 - \sqrt{var_{ck}} / 2 / \sigma ) ) )$.
The important property of this function is that it decreases as the width of the distribution increases above roughly $2 \sigma$.
Crossing below this value means that the variance of the prediction is approaching the ground truth variance we are regressing to.
This weight factor is then used to multiply the quadratic loss associated with this detection.

The equation for self-supervised keypoint estimation is obtained by solving the weighted least-squares problem discussed in \sref{sec:quadratic_magic}.
If we define the location of camera $c$ as $\mathbf{a}_c$ and the normalized direction of the ray from camera $c$ through keypoint $k$ as $\mathbf{\hat{d}}_{ck}$, the following equation yields the estimated 3D keypoint location:

\eqshrink
\begin{equation}
    \widetilde{\mathbf{x}}_k = \left(\sum_c^C \mathbf{I} - \mathbf{\hat{d}}_{ck} \mathbf{\hat{d}}_{ck}^\intercal \right )^{-1} \left ( \sum_c^C w_{ck} (I - \mathbf{\hat{d}}_{ck} \mathbf{\hat{d}}_{ck}^\intercal) \mathbf{a}_c \right ) 
    \label{eq:weighted_unsup_solution}
\end{equation}
\eqshrinkb

This is treated as a label and is not back propagated through.
The self-supervised loss is defined as: 

\eqshrink
\begin{equation}
    \mathcal{L}_{unsup} = \sum_{c}^{C} \sum_{k}^{K} D_\mathrm{KL} \left (\widetilde{H}_{ck} \middle|\middle| softmax(f(I_c, \theta)_{k}) \right )
\end{equation}
\eqshrinkb

Where $\widetilde{H}_{ck}$ is created from our estimated keypoints locations $\widetilde{\mathbf{x}}_k$ rather the annotated locations $\mathbf{x}_k$.
In the final loss we also include L2 regularization on all non-bias parameters.

\eqshrink
\begin{equation}
    \mathcal{L}_{total} = \mathcal{L}_{sup} + \alpha \mathcal{L}_{unsup} + \lambda || \theta_{non bias} ||^2
\end{equation}
\eqshrinkb

\subsection{Network architecture}

\begin{wrapfigure}{R}{7cm}
    \vspace{-1.0cm}
    \includegraphics[width=\linewidth]{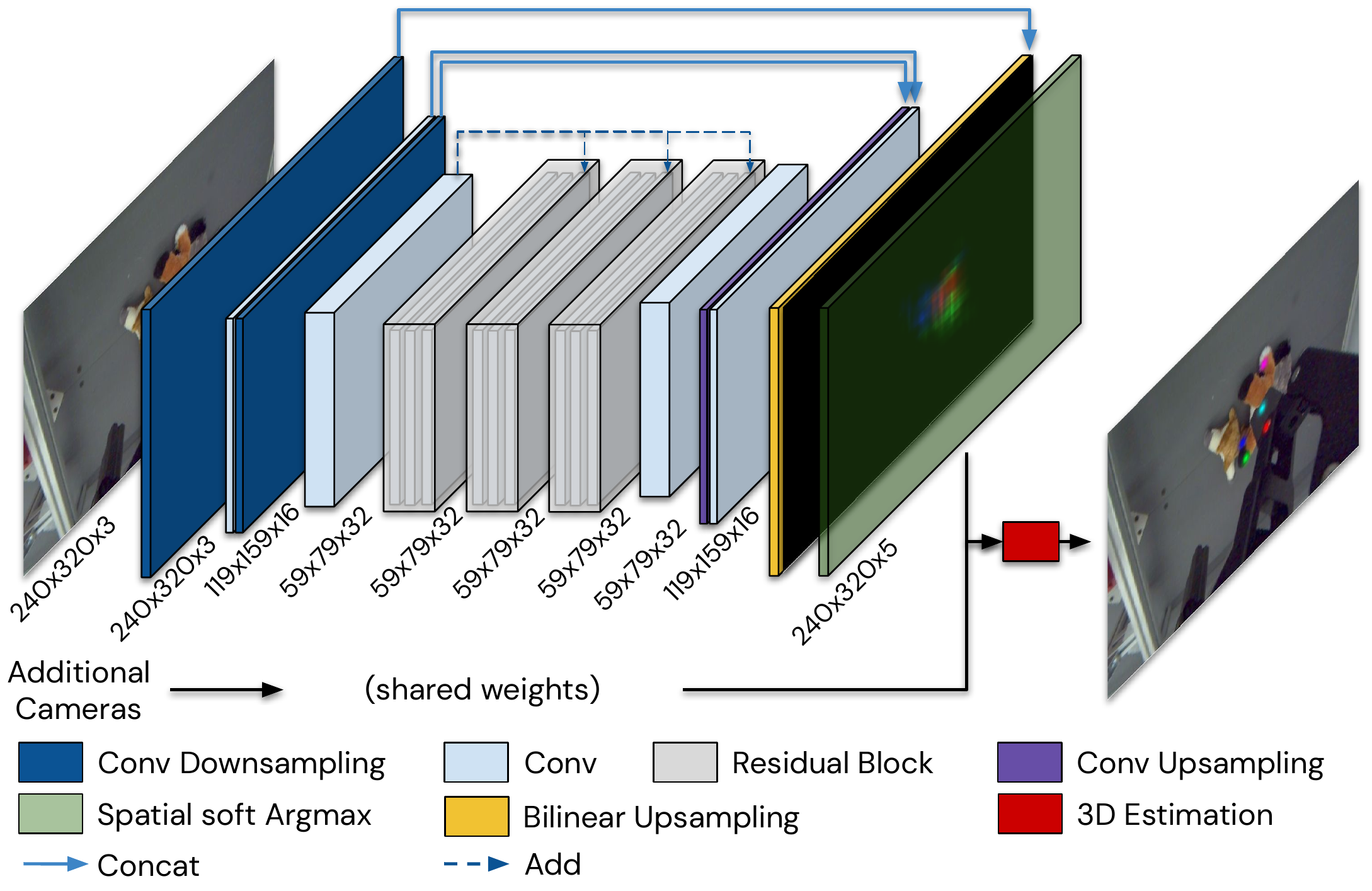}
    \caption{Network architecture used.}
    \label{fig:diag:unet}
    \vspace{-0.4cm}
\end{wrapfigure}

We follow standard practice in using a convolutional network to process images and obtain high-level features.
However, fundamentally we care about pixel-accurate detection -- \emph{where} these keypoints are in the image.
U-net's architecture \cite{ronneberger2015_unet} is a good fit for this problem, by allowing both a deep convolutional downsampling path for large receptive field as well as a resolution-preserving path for precision.

The input to the model is a single camera image and the output is a heat-map over all keypoints from that point of view.
Since we have multiple cameras providing images at each timestep we apply the model with the same weights to all camera views.
The network diagram is presented in \fref{fig:diag:unet}.

On the lowest resolution we apply blocks inspired by ResNet\cite{He2015Residual} with $3$ convolutional layers and $32$ channels in the intermediate layers.
We use a relatively small number of channels ($16$ to $32$), as using more did not increase accuracy, but decreased the framerate below our requirements for control.

\section{Investigating the model}
\label{sec:exploring_model}

In our experiments we aim to answer the following questions:
(1) What is the relative value of labelled vs. unlabelled data?
(2) Can the self-supervision mechanism be used to provide robustness to domain shift in the absence of further labels?
(3) Is the model sufficiently accurate and stable to use for real-robot control, and how does this compare to a non-geometric alternative?

\subsection{Value of labelled vs. unlabelled data}
Unlabelled data is typically less useful for training than labeled data, so one of our key objectives is to evaluate the extent to which this holds for S3K.
For this experiment we use a simulation, because it allows us to generate large amounts of well controlled data.

\textbf{Dataset.}
\label{sec:simple_sim_dataset}
To gather data we created a simple simulated environment similar to our physical robotic cell using MuJoCo\cite{todorov2012mujoco} and added a realistic 3D scanned box of a boardgame.
The keypoints were annotated at 3 corners of the box. 
For each sample we randomized the box location and orientation, as well as directions of 4 cameras from which we render the scene.
All images were $400$px by $400$px.
We generated a set of datasets of labelled and unlabelled data of various sizes.
The unlabelled datasets included the camera positions, but not the keypoint positions.
For each experiment we used one labelled and one unlabelled dataset and trained the model for 150k batches.
RMS pixel error on a withheld set of samples is used as the evaluation metric and is averaged over 3 seeds.

\begin{figure}[!tbp]
  \centering
  \begin{minipage}[t]{0.48\textwidth}
    \includegraphics[width=\textwidth]{figures/sup_unsup_2.pdf}
    \caption{
    Comparing effects of dataset sizes on keypoint detection error on a simulated task.
    Each curve corresponds to a number of \textit{unlabelled} datapoints.
    }
    \label{fig:fig:supunsup}
  \end{minipage}
  \hfill
  \begin{minipage}[t]{0.48\textwidth}
    \includegraphics[width=\textwidth]{figures/sup_unsup_3.pdf}
    \caption{
    Scaling of prediction error with total amount of data.
    Color denotes the ratio between the labelled and unlabelled data.
    }
    \label{fig:fig:supunsup_colors}
  \end{minipage}
\vspace{-0.2cm}
\end{figure}

\textbf{Results and discussion.}
\fref{fig:fig:supunsup} shows that with 3 labelled samples the model cannot successfully learn, but that 10 labelled samples (\ie 40 images) is sufficient to obtain a good model if given enough unlabelled data.
There is also a steady improvement in the performance as the amount of unlabelled data increases, especially in cases where little labelled data is provided.

Another way to visualize these results is to plot the total amount of data on the x-axis and use color to denote the ratio between the amounts of labelled and unlabelled data as in \fref{fig:fig:supunsup_colors}.
Here we see a clear trend in accuracy as a function of data, \textit{regardless of the kind of data}. 
This result suggests that beyond a small threshold, labelled and unlabelled data are equally valuable.
A careful observer may notice that almost fully labelled datasets perform worse than mixed ones.
This is an artifact of the training which arises because the two losses are combined with $\alpha = 0.5$ and the model overfits to the unlabelled data.
\label{sec:unstable_low_unsup}

\subsection{Domain shift experiments}
\label{sec:domain_shift}

The scenario that we are exploring here is one in which the environment distribution changes after the robot was deployed.
To simulate this we assume that we have annotations of an initial scene.
This scene consists of our object of interest as well some other task-irrelevant distractor objects.
The distractor objects change over time, but only unlabelled data from this process is available.
After the distractors are switched multiple times, we evaluate the ability of the model to predict the keypoint locations with new distractor objects present.

\textbf{Dataset.}
This dataset is based on the one described in \sref{sec:simple_sim_dataset} with extra objects of similar size and textures.
We call the scenario described above the \emph{start mode} annotation.
Our baseline is comparing to a model which has access to labelled data spread throughout the whole timeline instead of just the beginning - we call this the \emph{full mode} annotation.

\begin{figure}[!tbp]
\centering
\begin{minipage}[c]{0.48\textwidth}
    \includegraphics[width=\textwidth]{figures/domain_shift_2.pdf}
    \caption{Using unsupervised data to gain robustness to domain shift.}
    \label{fig:fig:domain_shift}    
\end{minipage}
\hfill
\begin{minipage}[c]{0.48\textwidth}
    \includegraphics[width=0.93\textwidth]{figures/vae_keypoint_4.pdf}
    \caption{
    Success rates of agents on an audio cable insertion task comparing using different vision features for RL.
    }
    \label{fig:rl}
\end{minipage}
\vspace{-0.5cm}
\end{figure}

\textbf{Results and discussion.}
\fref{fig:fig:domain_shift} shows that having access to the \emph{full} data is beneficial.
This gap consistently narrows as we increase the amount of unlabelled data.
Also we can see that the \emph{critical number} of labelled data has shifted to about $50$ for this harder dataset.

We can draw conclusions from this experiment by considering \fref{fig:fig:domain_shift}, which looks at the case of changing the amount of labelled data for both \emph{full mode} and \emph{start mode} distributions.
Each line represents a fixed amount of unlabelled data.
We see that when we have little unlabelled data (\ie blue and orange) there is a big performance gap between the \emph{start} and \emph{full} versions especially as the amount of labelled data increases towards the right.
This means that the model is significantly overfitting to the initial distractors.
We can contrast this to the behaviour of the purple curve which represents having much more unlabelled data.
Here the gap between \emph{start} and \emph{full} is minimal showing that the model is much less sensitive to the choice of distribution for the labelled data.
This means that the unlabelled data can be used to address the issue of domain shift.

\begin{wraptable}{R}{3.5cm}
    \centering
    \vspace{-0.2cm}
    \begin{tabular}[t]{lcc}
      \hline
      &\textbf{Train} & \textbf{Test}\\
      \hline
      1  & 0.78 & 26.99\\
      2  & 1.59 & 14.65\\
      3  & 1.73 & 16.22\\
      6  & 1.65 & 15.05\\
      8  & 1.82 & 11.75\\
      12 & 1.74 & 10.77\\
      \hline
    \end{tabular}
    \caption{RMS pixel error on the shoe keypoints as a function of the number of unique training examples.}
    \label{tab:shoe_generalization}
    \vspace{-0.9cm}
\end{wraptable}

\subsection{Within class generalization}
\label{sec:shoe_generalization}

So far we have focused on tracking keypoints grounded on specific object instances.
However, nothing about this model or architecture precludes learning category-level semantics.
In this section we demonstrate that this model is indeed capable of tracking consistent parts of a single \textit{class} of objects.
For this purpose we created a set of datasets in which we annotated $16$ shoes at their \textit{front}, \textit{tongue} and \textit{heel}.
Examples of images from these datasets can be seen in \fref{fig:img:shoe}.

This time in addition to randomizing the object position at test-time, we also evaluate on 4 held-out shoes not seen during training.  
Building on the domain-shift experiment, we compare models that had access to ground-truth labels on varying numbers of shoes during training, from 1-12.   
In \tref{tab:shoe_generalization} we see that this model is able to generalize reasonably well to unseen shoes from as few as 2 training instances, and that accuracy increases with additional instances as expected.
Although the gap between training and test errors is large, in a real setting a 10px accuracy would be sufficient for a reliable grasp of either part of the shoe.
We encourage the reader to view the \href{https://sites.google.com/view/2020-s3k/home}{\color{blue}{supplementary video}} for qualitative results.

\section{Keypoints for scripted behaviours}
\label{sec:for_scripted}
Unlike unstructured visual representations, keypoints are human understandable and provide actionable representation for robotic controllers.
To demonstrate this, we put $3$ flexible plush toys into the robotic cell and trained the keypoint model to track the head, front paws, body and tail of a plush fox.
Pictures of this scene can be seen in \fref{fig:image:foxie}.
The other $2$ toys acted as distractors.
We arranged the toys into $10$ different arrangements and collected $3$k timesteps (i.e. $12.5$ minutes) from each.
We manually labelled a single frame in each arrangement and used the stationarity of the scene to propagate the labels into all of the timesteps similarly as in \cite{Manuelli2019kPAMKA}.
$7$ of these arrangements were used for training and $3$ for testing, which gave us $21$k training timesteps in total.
The model is able to track all of the keypoints well, except for the tip of the tail which varied significantly between views and wasn't always visible.
The front and back paws are also sometimes confused especially in scenarios where the head and tail are occluded by the gripper in multiple views.

To demonstrate the accuracy of the model we scripted a motor primitive that grasped at a specified 3D location.
We demonstrate this behavior by consistently grasping the fox by the front paw.
Note that rigid object 6D pose tracking would not be applicable here due to the deformability of the toy.
The direction of the grasp is deduced by considering the \emph{body} keypoint as well.
Since our keypoint is \emph{inside} the object there is no need for additional processing between the keypoint and grasp point as would be case if the keypoint was on the surface.
This behaviour was repeatedly run and the main failure point was when the right paw was not immediately graspable.
To achieve a more robust controller we could either improve the detections by collecting additional unlabelled data, or improve the controller itself by using the features for RL. 
Both of these are used in the next section.

This shows that this model is capable of generalizing from a small number of labelled scenes to novel arrangements.
The right paw is a small target within the workspace, yet it could be located reliably enough to perform the physical task.
This demonstrates a workflow where tasks requiring precision and generalization can be solved with only a small amount of human labour.

\section{Keypoints for reinforcement learning}
\label{sec:for_rl}

\begin{figure}
\centering
    \begin{minipage}{.48\textwidth}
        \includegraphics[width=\linewidth]{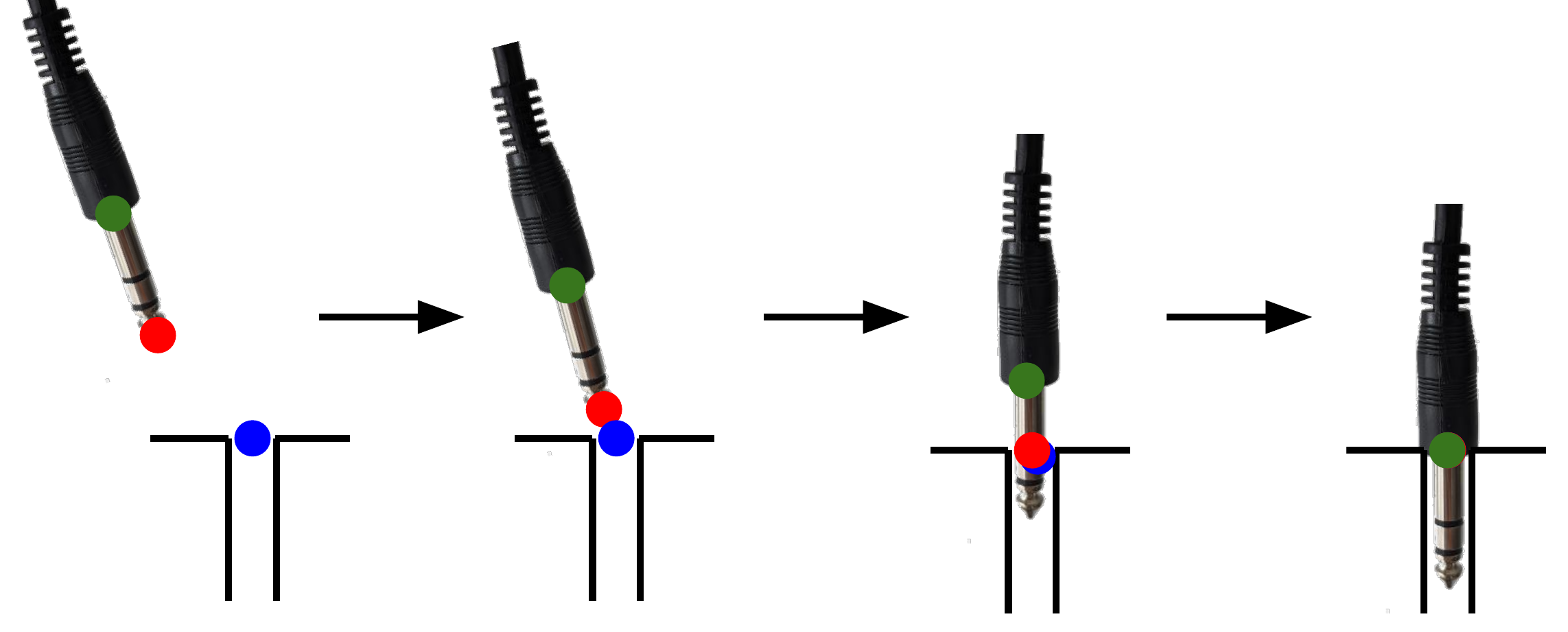}
        \captionof{figure}{Diagram of keypoints annotation on the cable.}
        \label{fig:diag:cable_annot}
    \end{minipage}%
    \hfill
    \begin{minipage}{.48\textwidth}
        \includegraphics[width=\linewidth]{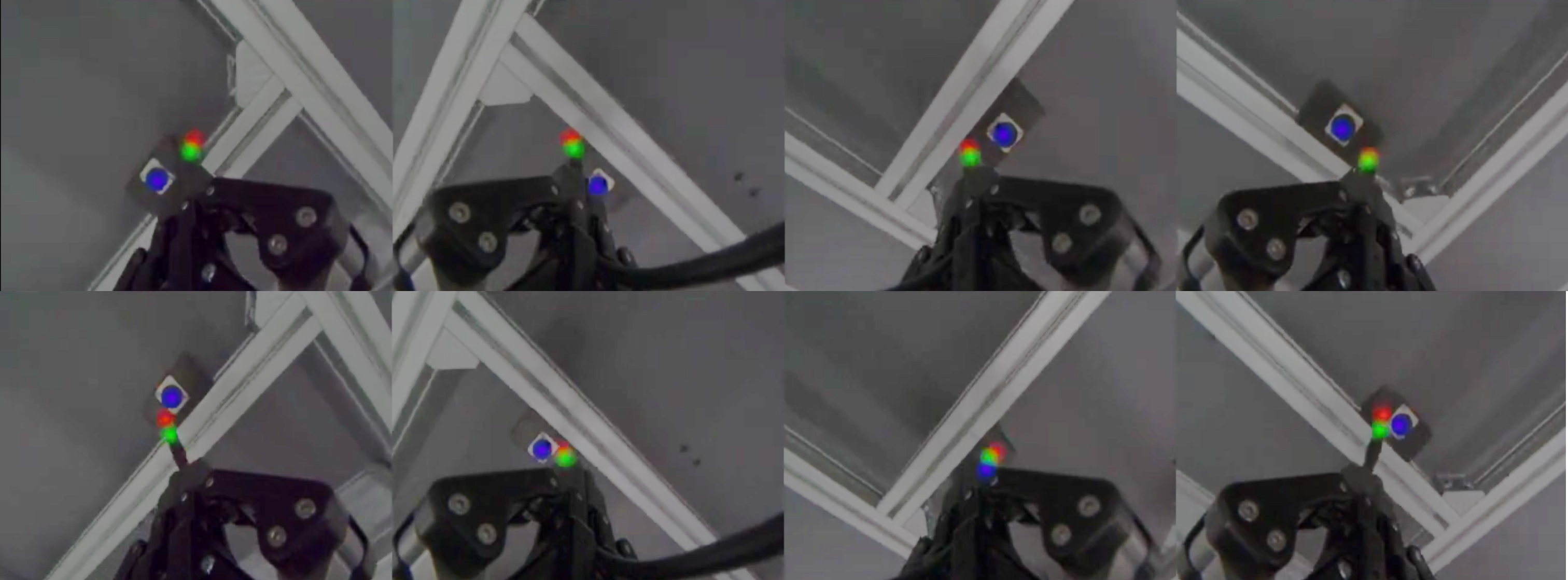}
        \captionof{figure}{
            Model detections for two points in time.
            Each row is a single time-step.
            }
        \label{fig:image:cable}
    \end{minipage}
\end{figure}

Our ultimate objective is to build agents that can be quickly taught to perform complex manipulation skills.  
So far we have focused purely on how the 3D keypoint representation allows for tracking and scripted control, but for our final experiment we sought to evaluate the representation on a task that would be difficult or impossible to script by hand.
Our main goal was to determine whether the 3D keypoint model would hold up in a closed loop RL setting, and if so, whether it could outperform a more general-purpose visual representation.

There are several reasons it might perform poorly, \eg instability in the model prediction, or bias to an overly restrictive feature-set.
However, by its nature RL can learn to be robust to multiple forms of noise, and strongly benefits from representations that allow generalization and data-efficiency.

\subsection{Task}
The task we selected was a deformable cable insertion problem, depicted in \fref{fig:image:cable_foxie_scenes}.  In this task the robot is holding a connector by the cable, and the goal is to insert it into a vertical socket.
The cable has sufficient flex to easily bend under gravity alone, and does not straighten to a consistent shape when removed. 
To randomize the environment between episodes we deliberately bend the cable using a scripted motion.
Actions were 6-dimensional Cartesian velocity commands at 4Hz and observations included gripper pose and velocity in addition to visual features.

This task is challenging to script because the cable deforms in complex ways when making contact with objects, and the relationship between robot actions and the cable state is highly non-linear.
Clearly two keypoints are insufficient to describe the full state of the cable, so this task also helps to evaluate whether the benefit of incorporating 3D structure is worth the cost of restricting the agent's attention to a finite number of points.
If the cable contour or other scene information is important, one might expect a VAE to outperform a keypoint model on this task.

\subsection{Agent}
We use a setup similar to EDRIAD \cite{Vecerk2019APA}, \ie a DPG \cite{silver2014deterministic} agent with demonstrations and keypoint locations as the visual features.
We used 51 successful episodes as demonstrations to accelerate training.
As keypoints we annotated the plug tip, plug base and socket positions as shown in \fref{fig:diag:cable_annot}.
As the plug tip inserts into the socket, we start tracking the socket opening instead, as we cannot consistently track the tip anymore\footnote{As discussed in \sref{sec:method}, the choice of what to track is implied by the labels -- \ie we labeled the plug-tip point at the socket opening after insertion.}.
This exploits an important feature of keypoints which is that they do not need to be attached to a specific point on the surface or a physical location as long as they are multi-view consistent.

As a baseline we trained an equivalent agent with a $\beta$-VAE\cite{Higgins2017betaVAELB} as the vision model. 
It embedded all 4 camera views into a single 10 dimensional latent embedding.
Both models were trained on the same unlabelled data, but the keypoint model received extra supervision from the labels.
Even this baseline agent used keypoint-distance between the plug and socket for rewards, although any reward mechanism e.g. electrical-connectivity could be substituted.

We gathered $\approx 3$k timesteps for both vision models.
For the keypoint model we labelled $120$ of them for the training and additional 36 for testing to enable hyperparameter tuning using vizier \cite{vizier}.
We defined the task success as the moment when the sum of the distances between all $3$ keypoints is below $2.5$cm as this means that the plug is almost fully inserted.
This distance minimizes false-negatives while preventing premature termination.
For safety, the episodes are also terminated if the agent moves far away from the socket or the episode takes too long (over 10s).

\subsection{Results and discussion}
Each experiment was run for 40000 time-steps which corresponds to about 3 hours of training.
We ran 5 runs for each setup and averaged the results.
Within each run we computed a rolling average over $2000$ timesteps.
As we can see in \fref{fig:rl}, the keypoint-based agent significantly outperformed ones trained with VAE features.
Error bars show min and max across seeds after the rolling average.
By the end of the experiment the keypoint agents reliably and repeatedly attempt insertion and final performance increases (above 95\%) correspond to their ability to succeed within the time limit.
VAE based agents never reach this performance even when trained for a significantly longer time as their training slows down after they reach success rate above 60\%.
We believe this is because the VAE representation lacks \emph{the precision} to master this task.
This shows that the keypoint representation is superior for these use-cases.

During the agent training, we would occasionally see keypoint mis-detections, however the chance of all keypoints collapsing to a small range was sufficiently small that we did not observe any false-positives.
This improves upon previous work where an ensemble had to be used to alleviate the issue of false positives \cite{Vecerk2019APA}.
These agents are capable of handling the noise from the model as well as implicitly reason about the deformations through the keypoint locations.
These results show that keypoints can act as a good \emph{summary} of the visual scene, and can be used as visual features for robotic tasks to avoid having to learn from raw images. 

\section{Conclusions}
\label{sec:conslusion}

In this paper we have built upon previous work in representations for reinforcement learning \cite{Vecerk2019APA}, supervised keypoint learning \cite{liu2020keypose} and geometry-based unsupervised keypoint learning \cite{rhodin2018unsupervised}.
We have introduced a new self-supervised loss that, when combined with a supervised loss on a small number of labelled samples, can provide a robust detector for semantic 3D keypoints.
We call this setup S3K and its main contribution comes from considering multi-view geometry as a source of self-supervision for keypoint based models and showing its applicability to robotic tasks.

We have conducted experiments on simulated data to investigate how S3K scales with different amounts of labelled and unlabelled data.
In particular, we showed that, given a small but sufficient set of labelled examples, further performance scales with the total amount of data, not the total labelled data.
We have further demonstrated how unlabelled data combined with S3K can be used to counteract effects of domain shift and shown its ability to generalize across samples from the same category.

To demonstrate applicability on real robot scenarios we  used the model outputs to script a policy to lift a plush toy fox by its front right paw in the presence of distractor objects.
In addition we apply S3K to reinforcement learning by using it to define a reward function and to provide visual features for an agent to learn a policy to insert a flexible audio cable into a socket.

We believe that our self-supervised keypoint detection via multiview consistency could be applied to many related fields as a way to reduce supervision.
We can see a trend in the field as we move away from supervised models towards models which require less and less supervision.
S3K dramatically reduces the need for supervision, but we believe that in the future we will be able to develop related methods which will remove the need for any explicit human grounding in terms of keypoints completely, but instead will be supervised purely via the task or demonstrations.

\clearpage

\section{Acknowledgments}
We would like to thank Giulia Vezzani, Michael Bloesch, and Martin Riedmiller for helpful reviews as well as Serkan Cabi, Misha Denil for fruitful discussions.
Further we would like to thank Francesco Nori for making this work possible.
We would also like to thank Federico Casarini, Stefano Salicety and Nathan Batchelor for help with the physical setup of the cell and Toby Turner and Dante Pomells for helping us setup the experiments when we couldn't be physically present.

\bibliography{main}  

\clearpage

\begin{appendices}

\section{Hardware details}

In all physical experiments we used 4 cameras.
They were RGB Basler daA1280-54ucm S-Mount with Evetar Lens M118B029528W F2.8 f2.95mm 1/1.8".
The robot was a Sawyer with a 2f-85 Robotiq gripper and FT 300 Force Torque Sensor which was used for a velocity admittance control to ensure the safety of the system.
We used custom 3D printed fingers to grasp the cable in a stable way.

\section{Simulated environment details}

All of our experiments used pixel error as the main metric.
We would like to provide some further information to allow a more intuitive understanding.
In \sref{sec:simple_sim_dataset} 1 pixel corresponded to about 2.8mm in the middle of the workspace.
Therefore 30px are about 8.4cm.
The size of the box was 19.9x1.7x10.5cm.

The provided measurements above apply to the experiments in \sref{sec:domain_shift} as well.
10px corresponds to 2.8cm which is relatively small compared to the overall size of the box.
The shoes in section \sref{sec:shoe_generalization} have realistic sizes, \ie about 25-30cm long.
Images of these shoes can be seen in \fref{fig:image:shoes_white}.

\section{Choosing weight for self-supervised loss}

In \sref{sec:unstable_low_unsup} we saw instabilities for small amounts of unsupervised data.
Therefore we explored how these effects change with the relative weight $\alpha$ between the supervised and unsupervised losses.
We used the domain shift dataset from \sref{sec:domain_shift} in the \emph{full} mode with $10$k unsupervised data points.

\begin{figure}[h]
\centering
    \begin{minipage}{.48\textwidth}
        \includegraphics[width=\linewidth]{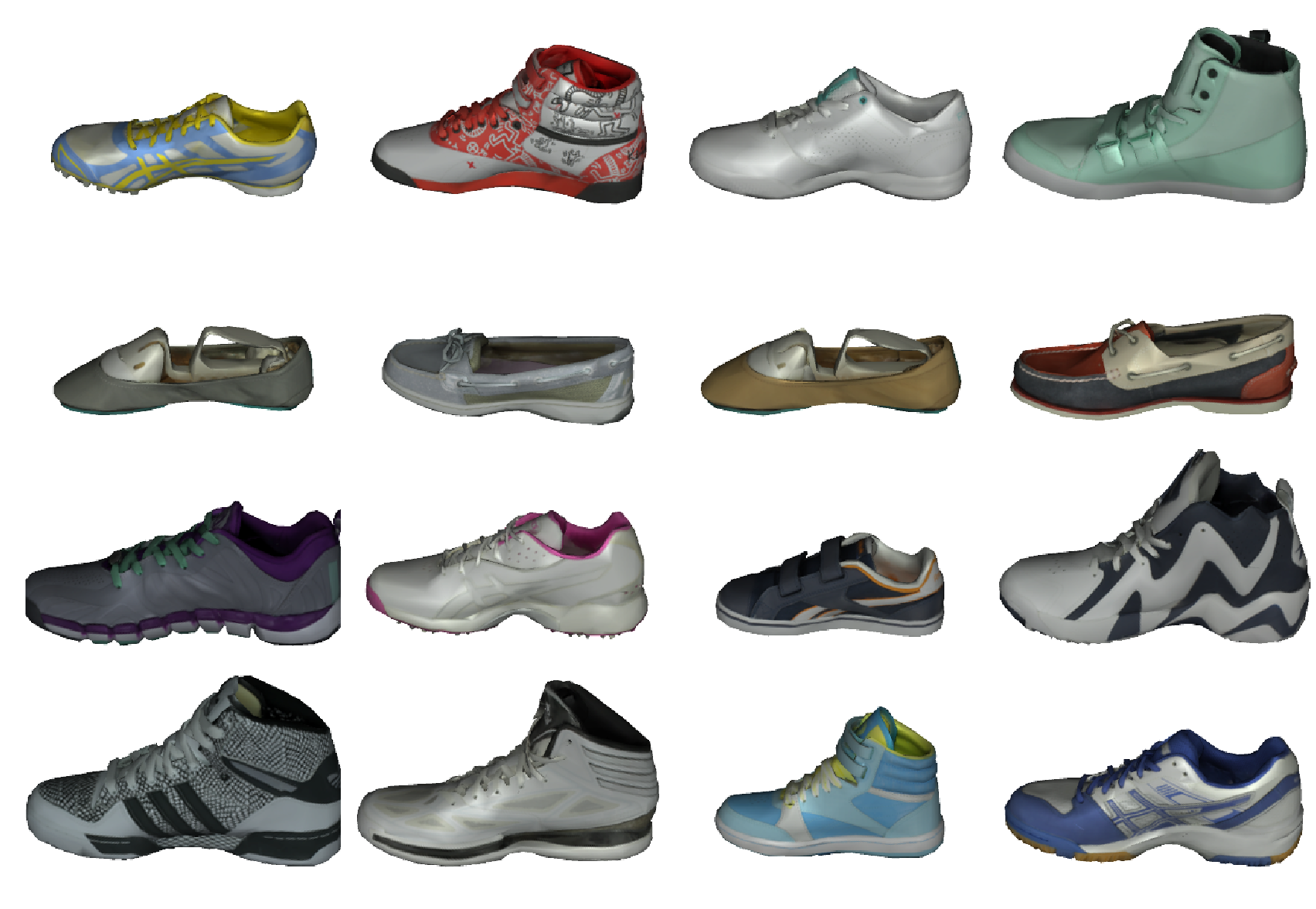}
        \caption{
        Shoes used in the generalization experiments.
        Top 3 rows were used in training and bottom row was used for evaluation.
        }
        \label{fig:image:shoes_white}
    \end{minipage}%
    \hfill
    \begin{minipage}{.48\textwidth}
        \includegraphics[width=\linewidth]{figures/unsup_loss.pdf}
        \caption{Choosing unsupervised loss weight $\alpha$.}
        \label{fig:fig:unsup_loss}
    \end{minipage}
\end{figure}

In \fref{fig:fig:unsup_loss} we see that the training can exhibit 2 main behaviours. 
If we don't provide enough labelled data the model collapses to a trivial solution and is unable to use the unlabelled data well.
On the other hand with enough labelled data the model performs well with a wide range of values of $\alpha$, but works best with $\alpha = 1$, \ie large weight on the unsupervised dataset.
This however is not true for the small amounts of labelled data where decreasing the weight of the self-supervised loss helps to stabilize the training.
Lowering $\alpha$ prevents the model from collapsing on simple 3D consistent solutions such as tracking the centre of the box which would be preferred by the self-supervised loss in the absence of labels.
Overall as we decrease the amount of unlabelled data we see that the optimal unsupervised loss gradually decreases.
In other experiments we used a weight of $\alpha = 0.5$.

\end{appendices}

\end{document}